\documentclass[a4paper]{svproc}
\usepackage{graphicx}
\usepackage{multicol}
\usepackage{footmisc}
\usepackage{color}
\usepackage{cite}
\usepackage{hyperref}
\usepackage{blindtext,graphicx}
\usepackage[absolute]{textpos}
\definecolor{somegray}{rgb}{0.5, 0.5, 0.5}
\newcommand{\darkgrayed}[1]{\textcolor{somegray}{#1}}
\makeatletter
\newcommand*\titleheader[1]{\gdef\@titleheader{#1}}
\AtBeginDocument{%
  \let\st@red@title\@title
  \def\@title{%
    \vskip-3em
    \bgroup\normalfont\large\centering\@titleheader\par\egroup
    \vskip1.1em\st@red@title}
}

\makeatother

\titleheader{\darkgrayed{This paper has been accepted for publication at the \\ IEEE International Conference on Robotics and Automation (ICRA), Philadelphia, 2022. \copyright IEEE}}

\TPGrid[1mm,1mm]{10}{15} 

\begin{document}
\mainmatter              

\begin{textblock}{8}(1,1)
\noindent\large\centering \darkgrayed{This paper has been accepted for publication at the \\ International Symposium on
Robotics Research (ISRR), Geneva, 2022.}
\end{textblock}
%
\title{Learning Agile, Vision-based Drone Flight: \\from Simulation to Reality}
%
\titlerunning{Learning Agile, Vision-based Drone Flight: from Simulation to Reality}  
%
\author{Davide Scaramuzza\inst{1} \and Elia Kaufmann\inst{1}}
\authorrunning{Davide Scaramuzza et al.} 
%
\tocauthor{Davide Scaramuzza, Elia Kaufmann}
\institute{University of Zurich}

\maketitle              

\begin{abstract}

We present our latest research in learning deep sensorimotor policies for agile, vision-based quadrotor flight.
We show methodologies for the successful transfer of such policies from simulation to the real world.
In addition, we discuss the open research questions that still need to be answered to improve the agility and robustness of autonomous drones toward human-pilot performance.
\end{abstract}

\section{Introduction}
Quadrotors are among the most agile and dynamic machines ever created. However, developing fully autonomous quadrotors that can approach or even outperform the agility of birds or human drone pilots with only onboard sensing and computing is very challenging and still unsolved.
Prior work on autonomous quadrotor navigation has approached the challenge of vision-based autonomous navigation by separating the system into a sequence of independent compute modules~\cite{Cieslewsk17iros, Ryll19icra, Mahony12ram,Shen2013aggressive}. 
While such modularization of the system is beneficial in terms of interpretability and allows to easily exchange modules, it results in a substantial increase in latency from perception to action and results in errors being propagated from one module to the next. 
Furthermore, prior work on autonomous quadrotor navigation heavily relies on control techniques such as PID control~\cite{Falanga17icra,Faessler18ral, Sun18ral} or model predictive control~(MPC)~\cite{Falanga18iros, Nan22ral}. 
While these methods have demonstrated impressive feats in controlled environments~\cite{Falanga17icra, Foehn21scienceTime,Nan22ral, Sun18ral}, they require substantial amount of tuning, and are difficult, if not impossible, to scale to complex dynamics models without large penalties in computation time.

In this extended abstract, we summarize our latest research in learning deep sensorimotor policies for agile vision-based quadrotor flight.
Learning sensorimotor controllers represents a holistic approach that is more resilient to noisy sensory observations and imperfect world models.
Training robust policies requires however a large amount of data.
Nevertheless, we will show that simulation data, combined with randomization and abstraction of sensory observations, is enough to train policies that generalize to the real world.
Such policies enable autonomous quadrotors to fly faster and more agile than what was possible before with only onboard sensing and computation.
In addition, we discuss the open research questions that still need to be answered to improve the agility and robustness of autonomous drones toward human-pilot performance.

\begin{figure}[t]
    \centering
    \includegraphics[trim={0 5cm 0 0},clip,width=\columnwidth]{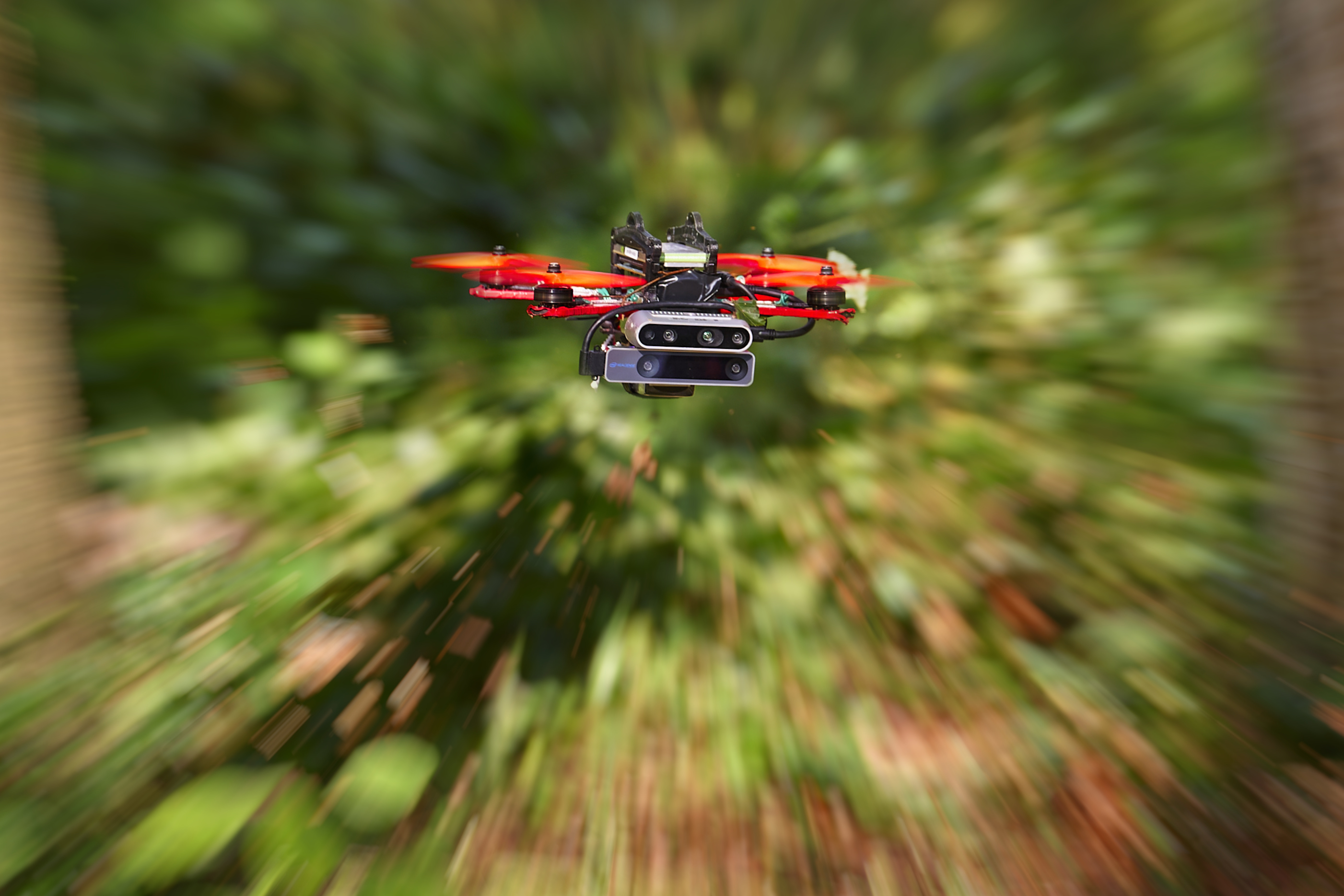}
    \caption{Our drone flying at high speed (10 m/s) through a dense forest using only onboard sensing and computation.}
    \label{fig:agile_flight}
\end{figure}

\section{High-Speed Flight in the Wild}

In~\cite{Loquercio21science}, we presented an end-to-end approach that can autonomously fly quadrotors through complex natural and human-made environments at high speeds (10 m/s), with purely onboard sensing and computation.\footnote{A video of the results can be found here: {\url{https://youtu.be/m89bNn6RFoQ}}}
The key principle is to directly map noisy sensory observations to collision-free trajectories in a receding-horizon fashion.
This direct mapping drastically reduces processing latency and increases robustness to noisy and incomplete perception.
The sensorimotor mapping is performed by a convolutional network that is trained \emph{exclusively} in simulation via privileged learning: imitating an expert with access to privileged information.
We leverage abstraction of the input data to transfer the policy from simulation to reality~\cite{Mueller18corl, Kaufmann20rss}.
To this end, we utilize a stereo matching algorithm to provide depth images as input to the policy.
We show that this representation is both rich enough to safely navigate through complex environments and abstract enough to bridge the gap between simulation and real world. 
By combining abstraction with simulating realistic sensor noise, our approach achieves zero-shot transfer from simulation to challenging real-world environments that were never experienced during training: dense forests, snow-covered terrain, derailed trains, and collapsed buildings.
Our work demonstrates that end-to-end policies trained in simulation enable high-speed autonomous flight through challenging environments, outperforming traditional obstacle avoidance pipelines.
A qualitative example of flight in the wild is shown in Figure~\ref{fig:agile_flight}.


%

\begin{figure}
    \centering
    \includegraphics[width=0.8\columnwidth]{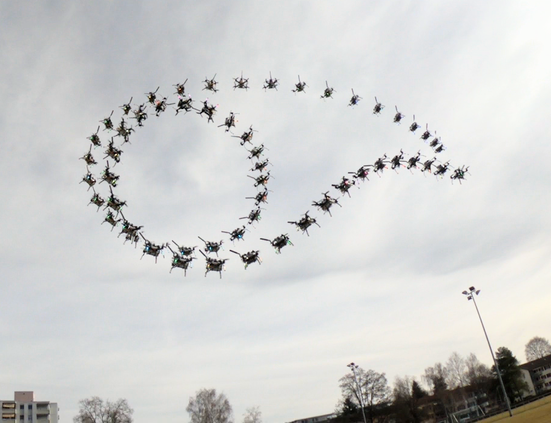}
    \vspace{0.5ex}
    \caption{Our quadrotor performs a Matty Flip. The drone is controlled by a deep sensorimotor policy and uses only onboard sensing and computation.}
    \label{fig:matty_loop}
\end{figure}
\section{Acrobatic Flight}

Acrobatic flight with quadrotors is extremely challenging.
Human drone pilots require many years of practice to safely
master maneuvers such as power loops and barrel rolls.
For aerial vehicles that rely only on onboard sensing and
computation, the high accelerations that are required for 
acrobatic maneuvers together with the unforgiving requirements
on the control stack raise fundamental questions in both
perception and control. 
For this reason, we challenged our drone with the task of performing acrobatic maneuvers~\cite{Kaufmann20rss}.\footnote{A video of the results can be found here: \url{https://youtu.be/2N_wKXQ6MXA}}
In order to achieve this task, we trained a neural network to predict actions directly from visual and inertial sensor observations.
Training is done by imitating an optimal controller with access to privileged information in the form of the exact robot state.
Since such information is not available in the physical world, we trained the neural network to predict actions instead from inertial and visual observations.

Similarly to the work described in the previous section, all of the training is done in simulation, without the need of any data from the real world. 
We achieved this by using abstraction of sensor measurements, which reduces the simulation-to-reality gap compared to feeding raw observations.
Both theoretically and experimentally, we have shown that abstraction strongly favours simulation-to-reality transfer.
The learned policy allowed our drone to go faster than ever before and successfully fly maneuvers with accelerations of up to 3g, such as the Matty flip illustrated in Figure~\ref{fig:matty_loop}.

\section{Autonomous Drone Racing}

Drone racing is an emerging sport where pilots race against each other with remote-controlled quadrotors while being provided a first-person-view~(FPV) video stream from a camera mounted on the drone.  
Drone pilots undergo years of training to master the skills involved in racing. 
In recent years, the task of autonomous drone racing has received substantial attention in the robotics community, which can mainly be attributed to two reasons:
(i)~The sensorimotor skills required for autonomous racing would also be valuable to autonomous systems in applications such as disaster response or structure inspection, where drones must be able to quickly and safely fly through complex dynamic environments.
(ii)~The task of autonomous racing provides an ideal test bed to objectively and quantifiably compare agile drone flight against human baselines. It also allows us to measure the research progress towards the ambitious goal of achieving super-human performance.

%

One approach to autonomous racing is to fly through the course by tracking a precomputed, potentially time-optimal, trajectory~\cite{Foehn21scienceTime}. 
However, such an approach requires to know the race-track layout in advance, along with highly accurate state estimation, which current methods are still not able to provide.
Indeed, visual inertial odometry  is subject to drift in estimation over time. 
SLAM methods can reduce drift by relocalizing in a previously-generated, globally-consistent
map. 
However, enforcing global consistency leads to increased computational demands that strain the limits of on-board processing.

Instead of relying on globally consistent state estimates, we deploy a convolutional neural network to identify the next location to fly to, also called waypoints.
However, it is not clear a priori what should be the representation of the next waypoint.
In our works, we have explored different solutions.
In our preliminary work, the neural network predicts a fixed distance location from the drone~\cite{Kaufmann18corl}.
Training was done by imitation learning on a globally optimal trajectory passing through all the gates.
Despite being very efficient and easy to develop, this approach cannot efficiently generalize between different track layouts, given the fact that the training data depends on a track-dependent global trajectory representing the optimal path through all gates.

For this reason, a follow-up version of this work proposed to use as waypoint the location of the next gate~\cite{Kaufmann19icra}.
As before, the prediction of the next gate is provided by a neural network.
However, in contrast to the previous work, the neural network also predicts a measure of uncertainty.

\begin{figure}
    \centering
    \includegraphics[width=\columnwidth]{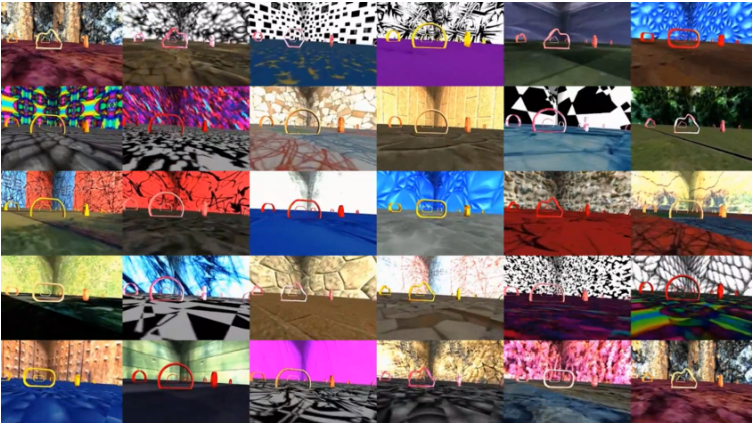}
    \vspace{0.2ex}
    \caption{The perception component of our system, represented by a convolutional neural network (CNN), is trained only with non-photorealistic simulation data.}
    \label{fig:domain_randomization}
\end{figure}

Even though the waypoint representation proposed in~\cite{Kaufmann19icra} allowed for efficient transfer between track layouts, it still required substantial amount of real-world data to train.  
Collecting such data is generally a very tedious and time consuming process, which represents a limitation of the two previous works.
In addition, when something changes in the environment, or the appearance of the gates changes substantially, the data collection process needs to be repeated from scratch.
For this reason, in our recent work~\cite{Loquercio19tro} we have proposed to collect data \emph{exclusively} in simulation.
To enable transfer between the real and the physical world, we randomized all the features which predictions should be robust against, \emph{i.e.} illumination, gate shape, floor texture, and background.
A sample of the training data generated by this process, called domain randomization, can be observed in Figure~\ref{fig:domain_randomization}.
Our approach was the first to demonstrate zero-shot sim-to-real transfer on the task of agile drone flight.
A collection of the ideas presented in the above works has been used by our team to successfully compete in the Alpha-Pilot competition~\cite{Foehn20rss}.\footnote{A video of the results can be found here: \url{https://youtu.be/DGjwm5PZQT8}}

\section{Future Work}

Our work shows that neural networks have a strong potential to control agile platforms like quadrotors.
In comparison to traditional methods, neural policies are more robust to noise in the observations and can deal with imperfection in sensing and actuation.
However, the approaches presented in this extended abstract fall still short of the performance of professional drone pilots. 
To further push the capabilities of autonomous drones, a specialization to the task is required, potentially through real-world adaptation and online learning. 
Solving those challenges could potentially bring autonomous drones closer in agility to human pilots and birds.
In this context, we present some limitations of the approaches proposed so far and interesting avenues for future work.

\subsection{Generalization}
One example of such a challenge is \emph{generalization}: even though this extended abstract presents methods presents promising approaches to push the frontiers of autonomous, vision-based agile drone navigation in scenarios such as acrobatic flight, agile flight in cluttered environments, and drone racing, it is not clear how these sensorimotor policies can be efficiently transferred to novel task formulations, sensor configurations, or physical platforms. 
Following the methodologies presented in this paper, transferring to any of these new scenarios would require to retrain the policy in simulation, or perform adaptation using learned residual models.
While the former suffers from the need to re-identify observation models and dynamics models, the latter is restricted to policy transfer between simulation and reality for the same task. 

Possible avenues for future work to address this challenge include hierarchical learning~\cite{haarnoja2018latent} or approaches that optimize policy parameters on a learned manifold~\cite{Peng20rss}. 

\subsection{Continual Learning}
The approaches to sensorimotor policy training presented in this paper are \emph{static} in their nature:
after the policy is trained in simulation via imitation learning or reinforcement learning, its parameters are frozen and applied on the real robot. 
In contrast, natural agents interact fundamentally different with their environment; they continually adapt to new experience and improve their performance in a much more dynamic fashion. 
Designing an artificial agent with similar capabilities would greatly increase the utility of robots in the real world and is an interesting direction for future work. 

Recent work has proposed methods to enable artificial agents to perform few-shot learning of new tasks and scenarios using techniques such as adaptive learning~\cite{Smith21legged, Kumar21rma} or meta learning~\cite{andrychowicz2016learning, finn2017model}.
These methods have shown promising results on simple manipulation and locomotion tasks but remain to be demonstrated on complex navigation tasks such as high-speed drone flight in the real world. 

\subsection{Autonomous Drone Racing}
The approaches presented in this paper addressing the challenge of autonomous drone racing are limited to single-agent time trial races. 
To achieve true racing behaviour, these approaches need to be extended to a multi-agent setting, which raises novel challenges in perception, planning, and control. 
Regarding perception, multi-agent drone racing requires to detect opponent agents, which is a challenging task when navigating at high speeds and when onboard observations are subject to motion blur. 
The planning challenges arise from the need to design game-theoretic strategies~\cite{Spica18rss} that involve maneuvers such as strategic blocking, which are potentially not time-optimal but still dominant approaches when competing in a multi-agent setting. 
Additionally, the limited field of view of the onboard camera renders opponents potentially unobservable, which requires a mental model of the trajectory of opponents. 
Finally, racing simultaneously with other drones through a race track poses new challenges in modeling and control, as aerodynamic effects induced by other agents need to be accounted for.

We envision that many of these challenges can be addressed using deep reinforcement learning in combination with \emph{self-play}, where agents improve by competing against each other~\cite{baker2019emergent,silver2017mastering}.

\section{Acknowledgments}
This work was supported by the National Centre of Competence in Research (NCCR) Robotics through the Swiss National Science Foundation (SNSF) and the European Union’s Horizon 2020 Research and Innovation Program under grant agreement No. 871479 (AERIAL-CORE) and the European Research Council (ERC) under grant agreement No. 864042 (AGILEFLIGHT).

{
\bibliographystyle{IEEEtran}
\bibliography{references.bib}
}

\end{document}